\documentclass[letterpaper, 10 pt, conference]{ieeeconf}  % Comment this line out if 
\IEEEoverridecommandlockouts                              % This command is only 
\usepackage{hyperref}
\usepackage{graphicx}
\usepackage{booktabs}
\usepackage{multirow}
\usepackage{cuted}
\usepackage{capt-of}
\usepackage{amsmath}
\usepackage{amsfonts}
\usepackage{tabularx}
\usepackage{array}
\usepackage{float}
\newcolumntype{Y}{>{\centering\arraybackslash}X}

\title{\LARGE \bf
TacForeSight: Force-Guided Tactile World Model for 

Contact-Rich Manipulation
}
\author{
Yujie Zang$^{1,2,3*}$,
Yuhang Zheng$^{2*}$,
Xian Nie$^{1,3*}$,
Yupeng Zheng$^{1, 4}$,
Shuai Tian$^{4}$,
Songen Gu$^{5}$,\\
Chen Gao$^{2}$, 
Zining Wang$^{1}$,
Shuicheng Yan$^{2\dagger}$,
Wenchao Ding$^{1\dagger}$\\
$^{1}$ TARS Robotics \quad
$^{2}$ National University of Singapore \quad
$^{3}$ Shanghai Jiao Tong University \\
$^{4}$ Institute of Automation, Chinese Academy of Sciences \quad
$^{5}$ Fudan University \\
\textsuperscript{$\dagger$} Corresponding Author,
\textsuperscript{*} Equal Contribution \\
}
\begin{document}
\maketitle
\begin{strip}
    \centering
    \includegraphics[width=\textwidth]{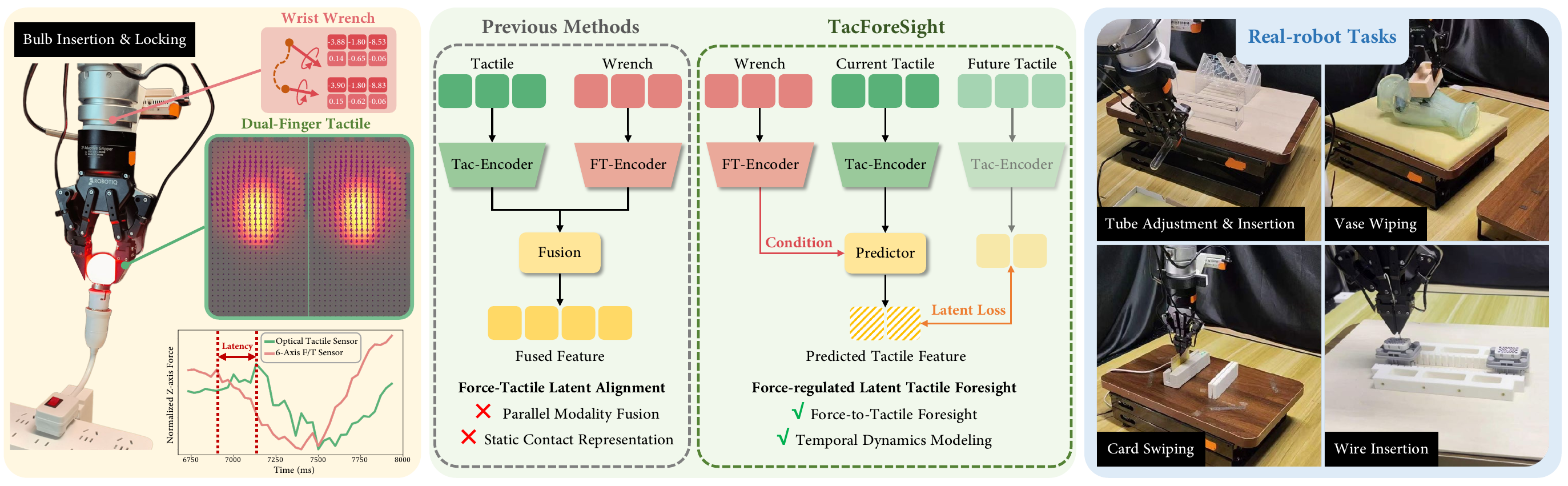}
    \captionof{figure}{
    \textbf{Left}: Wrist wrench and dual-finger tactile signals provide complementary global and local contact cues, with external force/torque variations preceding tactile responses during contact transitions.
    \textbf{Middle}: Unlike prior force--tactile fusion methods, TacForeSight predicts force-conditioned tactile evolution to provide anticipatory contact priors.
    \textbf{Right}: Real-robot evaluation across representative contact-rich manipulation tasks.
    }
    % \textbf{Left}: Wrist force/torque and dual-finger tactile signals provide complementary global and local contact information during bulb insertion and locking. 
    % \textbf{Middle}: Existing approaches typically fuse or align force and tactile features, whereas TacForeSight models force-conditioned tactile evolution to generate anticipatory tactile priors. 
    % \textbf{Right}: Real-robot evaluation across representative contact-rich manipulation tasks.
    % }
    \label{fig:teaser}
\end{strip}
\vspace{-15pt}

\thispagestyle{empty}
\pagestyle{empty}

% \begin{abstract}
\begin{abstract}
Contact-rich manipulation requires robots to continuously perceive and regulate evolving physical interactions under dynamic contact transitions or complex surface geometries. 
Recent imitation learning methods improve contact-aware control by incorporating tactile or force feedback, but they rarely model the asymmetric spatiotemporal roles of global force and local tactile sensing. 
To address this, we propose TacForeSight, a lightweight force-conditioned tactile foresight framework for real-time manipulation. 
The core component is TacForceWM, a tactile world model that predicts short-horizon tactile latent dynamics from dual-finger tactile observations conditioned on high-frequency wrist force and torque signals. 
Another key component, the Predictive Tactile-Conditioned Policy, leverages the predicted latents as anticipatory contact priors, models the current-to-future tactile evolution via cross-attention, and adaptively fuses visuo-tactile features through a tactile-guided gating module.
By forecasting purely within a compact latent space, TacForeSight enables proactive contact reasoning with efficient real-time inference suitable for high-frequency manipulation control.
Real-robot experiments on five representative tasks and three in-process perturbation settings show that TacForeSight consistently outperforms existing baselines, particularly under dynamic contact disturbances. 
All models and datasets will be made publicly available on the project website at \url{https://tacforesight.github.io/ProjectPage}.
% All models and datasets will be made publicly available at our
% \href{https://tacforesight.github.io/ProjectPage/}{project website}.

\end{abstract}

% \end{abstract}

\section{Introduction}
Contact-rich manipulation remains a fundamental challenge, as robots need to adapt to dynamic contact transitions or geometrically complex surfaces.
During execution, subtle changes in contact force, local geometry, or object pose can rapidly alter the interaction state, leading to slippage, misalignment, or contact loss. 
Successful execution therefore requires robots to continuously perceive and regulate evolving physical interactions between the robot and environment.

Recent imitation learning methods have improved contact-rich manipulation by incorporating tactile sensing or wrist force/torque feedback into visuomotor policies~\cite{li2026forcevla2, zheng2026omnivta}. 
However, most existing approaches either treat force and tactile signals as separate modalities to be fused into a joint representation~\cite{li2026master, feng2026anytouch, huang2026tactile}, or use them as passive feedback signals for reactive control~\cite{xue2025reactive, fang2026force}. 
Although these designs improve contact awareness, their inherently reactive nature limits their effectiveness in tasks that require proactive temporally coordinated interaction modeling. 
A key limitation of current methods is that they do not explicitly model the asymmetric temporal and spatial roles of global force and local tactile sensing. 
Wrist wrench signals provide high-frequency global cues about external load changes, whereas optical tactile sensors capture fine-grained local deformation. 
For example, in bulb insertion and locking (Fig.~\ref{fig:teaser}), abrupt wrist-wrench changes consistently precede fingertip tactile responses become evident, indicating a coarse-to-fine temporal dependency where global force signals act as leading indicators of future tactile states.
Effectively exploiting this relationship, however, requires moving beyond existing reactive feature fusion toward explicit modeling of cross-modal dynamics.

Inspired by human sensorimotor control, where load-related cues are used to anticipate object interaction states and guide local contact adjustments~\cite{johansson2009coding, flanagan2001sensorimotor}, we explicitly model this force-to-tactile predictive process for contact-rich manipulation. 
Therefore, we propose \textbf{TacForeSight}, a force-conditioned tactile prediction framework that predicts short-horizon future tactile representations from wrist force/torque dynamics and uses the predicted tactile to guide action prediction during upcoming contact transitions.

Specifically, \textbf{TacForeSight} consists of two key components.
The first is \textbf{TacForceWM}, a force-conditioned tactile world model that predicts short-horizon tactile dynamics within a latent space.
The second is a \textbf{Predictive Tactile-Conditioned Policy}, which leverages the predicted tactile latents as anticipatory contact priors for action sequence prediction.
Within the policy, a current–future tactile interaction module models temporal tactile evolution via cross-attention, while a lightweight visuo-tactile fusion module adaptively integrates multimodal observations.
Unlike existing frameworks~\cite{xu2025exumi, zheng2026omnivta} that rely on computationally intensive predictive branches or high-dimensional video generation, TacForeSight focuses on efficient latent-space forecasting. This enables proactive contact reasoning while maintaining a real-time inference rate of 20~Hz on an RTX 4090D GPU.

We evaluate the effectiveness of the TacForeSight on five representative contact-rich manipulation tasks and three in-process perturbation settings that explicitly disrupt ongoing contact states. TacForeSight achieves state-of-the-art performance across both nominal and perturbation settings, demonstrating that force-conditioned tactile foresight improves contact establishment, contact maintenance, and recovery from dynamic contact disturbances.

The main contributions of this paper are as follows:
\begin{itemize}
\item We propose \textbf{TacForeSight}, a compact force-conditioned tactile prediction framework for real-time contact-rich manipulation, which achieves superior robustness under various disturbances compared to relevant methods.

\item We introduce \textbf{TacForceWM}, a tactile world model that predicts short-horizon tactile latent dynamics from dual-finger tactile observations conditioned on high-frequency wrist force/torque signals.

\item We design a \textbf{Predictive Tactile-Conditioned Policy} that uses future tactile latents as anticipatory contact priors, models current--future tactile evolution via cross-attention, and performs adaptive visuo-tactile fusion with a tactile-guided gate.
\end{itemize}

\section{RELATED WORKS}
\subsection{Contact-rich Manipulation with Tactile and Force}
Recent studies have shown that incorporating force and tactile sensing can significantly improve the robustness and success rate of contact-rich manipulation. Prior works leverage force or tactile sensing for contact-aware planning and closed-loop correction. Methods such as FoAR~\cite{he2025foar}, TacVLA~\cite{zhang2026tacvla}, and OmniVTA~\cite{zheng2026omnivta} dynamically regulate visual and force/tactile modalities through gating mechanisms. Additionally, some methods~\cite{yu2026forcevla, li2026forcevla2, li2026favla, bi2025vla} further integrate wrench or tactile observations with the visual and language modalities, thereby improving the policy’s capability to model physical interaction dynamics. Other approaches~\cite{xue2025reactive, fang2026force, choi2026wild, li2026master} use force/tactile feedback for reactive closed-loop control. 
Despite these advances, existing methods mainly use force and tactile sensing as additional policy observations, either through direct fusion, adaptive gating, or shared latent alignment~\cite{feng2026anytouch, huang2026tactile, feng2025anytouch}. 
Such designs improve reactive contact feedback, but they rarely model the cross-modal dynamics between global force variations and local tactile state evolution. 
As a result, the predictive relationship between wrist force/torque and future tactile contact states remains underexplored.

In this paper, we propose TacForeSight. Different from prior approaches, our method introduces force conditioning into latent dynamics modeling to forecast future tactile representations conditioned on global wrist force/torque dynamics.

\subsection{World Model for Robotic Manipulation}
World models for robotic manipulation predict future states to support planning or condition policies on anticipated scene and physical evolution~\cite{wu2023daydreamer, kim2026cosmos, gu2026vistabot, liu2025mla}. 
Recent works mainly focus on visual or video prediction, generating future RGB frames or latent visual features for manipulation and action generation~\cite{kim2026cosmos, zhou2024robodreamer, zheng2025world4drive, li2026causal}. 
These models are also used as learned simulators or policy-conditioning modules, reflecting a shift from reactive perception to predictive control~\cite{liao2025genie, team2025evaluating}.
However, contact-rich manipulation requires anticipating not only visual changes, but also contact transitions, force variations, and local deformation. 
Some methods directly predict future force or tactile observations for planning~\cite{higuera2026visuo, zheng2026omnivta}, but generating high-dimensional sensory observations, especially in pixel space, is computationally expensive for real-time control~\cite{higuera2026visuo}. 
Other works use force or tactile prediction as an auxiliary objective for representation learning~\cite{xu2025exumi, ye2025learning}, without explicitly using the predicted contact information as prior during execution.

In contrast, TacForceWM predicts short-horizon tactile evolution in a compact latent space conditioned on high-bandwidth wrist force/torque signals.
The predicted tactile latents serve as anticipatory contact priors, enabling proactive control in robotic manipulation.
% \clearpage

\begin{figure*}[t]
    \centering
    \includegraphics[width=\textwidth]{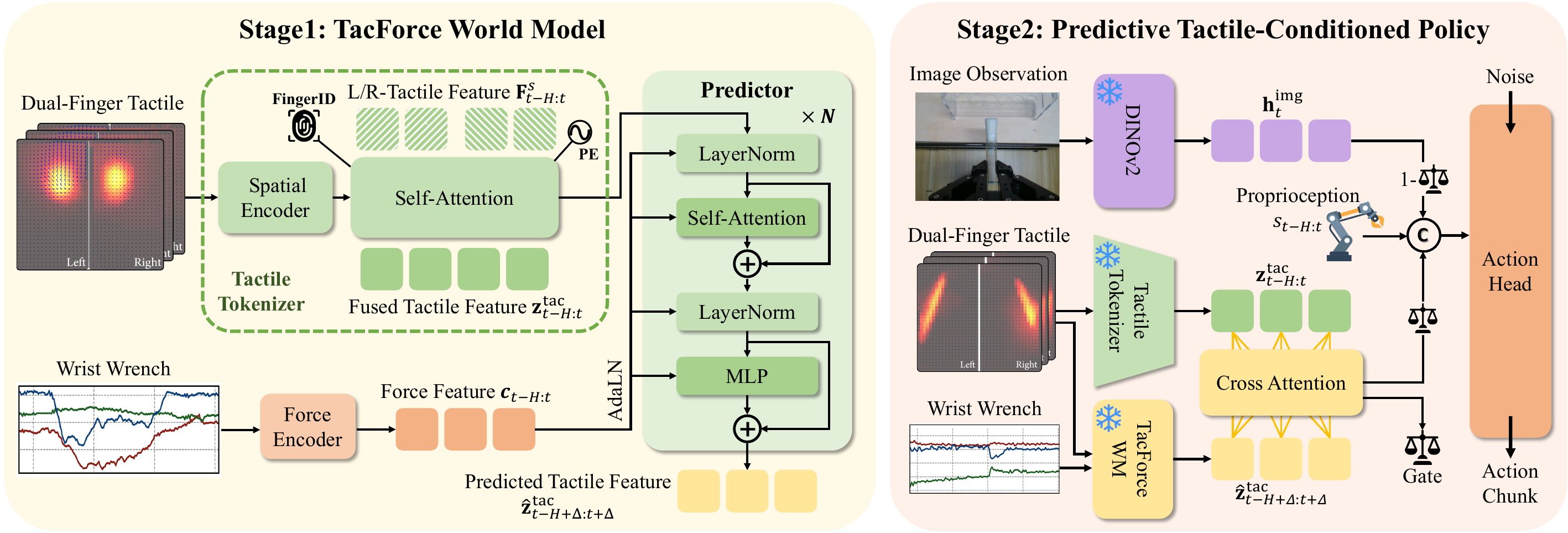}
    \vspace{-2mm}
    \caption{
    \textbf{Overview of TacForceSight.}
    Our framework consists of two coupled components. 
    In Stage 1, a force-conditioned tactile world model encodes dual-finger tactile fields into compact latent representations and predicts tactile evolution conditioned on wrist force/torque signals. 
    In Stage 2, the predicted tactile dynamics are used as contact priors for a lightweight flow-based policy. 
    }
    \label{fig:pipeline}
    \vspace{-5mm}
\end{figure*}

\section{METHOD}

To address the challenge of modeling physical interaction dynamics in contact-rich manipulation, we propose \textbf{TacForeSight}, a cascaded predictive framework that first pretrains a force-conditioned world model to predict short-horizon tactile evolution in latent space and then leverages the predicted tactile dynamics for lightweight flow-based action sequence prediction.

\subsection{Force-conditioned Tactile World Model}

Our tactile world model consists of a tactile tokenizer, a force encoder, and a latent dynamics predictor.
Given recent tactile observations and wrist wrench signals, it learns to model the force-conditioned evolution of tactile latent states during physical interactions.

\subsubsection{Tactile Tokenizer}
We implement the tactile tokenizer as a hybrid CNN-Transformer~\cite{dosovitskiy2020image} that converts dense dual-finger tactile fields into compact frame-level tokens.

Given a tactile observation at time $t$, each finger-specific tactile field is denoted as
$\mathbf{X}_t^s \in \mathbb{R}^{H \times W \times 3}$, where $s \in \{L,R\}$ indexes the left or right finger, $H \times W$ is the tactile spatial resolution, and the three channels represent dense 3D marker displacement at each spatial location.
We first employ a shared spatial encoder $\Phi_{\mathrm{sp}}$, consisting of residual convolutional blocks and hierarchical downsampling layers, to extract local tactile deformation features:
\begin{equation}
    \mathbf{F}_t^s = \Phi_{\mathrm{sp}}(\mathbf{X}_t^s) \in \mathbb{R}^{H' \times W' \times D_h}.
\end{equation}
The encoder weights are shared across both fingers for dual-finger tactile observations~\cite{lee2026symmetry}.

To encode spatial contact geometry and distinguish finger-specific interaction roles, we augment the feature maps with learnable spatial positional embeddings $\mathbf{E}_{\mathrm{pos}}$ and finger-specific identity embeddings $\mathbf{E}_{\mathrm{id}}^s$:
\begin{equation}
    \tilde{\mathbf{F}}_t^s = \mathbf{F}_t^s + \mathbf{E}_{\mathrm{pos}} + \mathbf{E}_{\mathrm{id}}^s.
\end{equation}

The feature maps are flattened into patch-level tactile tokens and prepended with a learnable \texttt{[CLS]} token. 
A Transformer is applied to the concatenated left and right finger tokens, enabling allowing self-attention to capture spatial contact patterns within each finger as well as interaction dependencies across fingers.  The \texttt{[CLS]} output is taken as the frame-level tactile latent
$\mathbf{z}_t \in \mathbb{R}^{D_z}$.
This latent provides a compact representation of the joint in-hand interaction state for subsequent latent dynamics prediction.

\subsubsection{Force Encoder}
To incorporate global physical interaction cues, we design a temporal force encoder that extracts temporally consistent conditions from high-rate wrist force/torque signals. 
Given a six-axis force/torque sequence $\mathbf{w}_{t-nH:t} \in \mathbb{R}^{nH \times 6}$ sampled at a higher rate than tactile observations, where $n$ denotes the sampling-rate ratio, the encoder maps it into a tactile-aligned condition sequence:
\begin{equation}
    \mathbf{c}_{t-H:t} = G_{\phi}(\mathbf{w}_{t-nH:t}).
\end{equation}

The encoder first projects raw force/torque vectors into a latent feature space, and then applies dilated causal 1D convolutional blocks to capture multi-scale temporal force variations~\cite{van2016wavenet}. 
A causal temporal downsampling layer aligns the high-rate force features with the tactile latent sequence. 
This design preserves temporal causality while converting continuous wrist wrench patterns into physically informative conditions for latent tactile dynamics prediction.

\subsubsection{Latent Dynamics Predictor}
This component explicitly conditions spatial tactile prediction on temporal force features to capture highly dynamic physical interactions.
To keep prediction fast and lightweight, we adopt a latent predictive formulation that forecasts future tactile representations rather than reconstructing raw tactile observations~\cite{bardes2023v}.
Instead of performing frame-wise one-step prediction~\cite{maes2026leworldmodel}, we formulate the task as chunk-based forecasting to further obtain temporally coherent dynamics. Concretely, given a tactile latent chunk $\mathbf{z}_{t-H:t}$ and its aligned force condition $\mathbf{c}_{t-H:t}$, the predictor estimates a future tactile latent chunk with temporal offset $\Delta$:
\begin{equation}
    \hat{\mathbf{z}}_{t-H+\Delta:t+\Delta}
    =
        T_{\psi}
        \left(
            \mathbf{z}_{t-H:t}^\mathrm{{tac}},
            \mathbf{c}_{t-H:t}^\mathrm{{tac}}
        \right),
\end{equation}
where $T_{\psi}$ denotes a force-conditioned latent Transformer backbone.
The force condition is injected through adaptive layer normalization (AdaLN)~\cite{peebles2023scalable} to modulate the intermediate features according to the temporal wrench context.

\subsubsection{Training Objectives}
The world model is trained with a prediction objective and a latent regularization objective.

The prediction objective jointly supervises absolute future tactile latents and their first-order temporal dynamics to capture tactile evolution and reduce over-smoothed latent predictions.
Let
$\mathbf{Z}_{t}^\mathrm{{tac}} = \mathbf{z}_{t-H+\Delta:t+\Delta}^\mathrm{{tac}}$
and
$\hat{\mathbf{Z}}_{t}^\mathrm{{tac}} = \hat{\mathbf{z}}_{t-H+\Delta:t-H+\Delta}^\mathrm{{tac}}$
denote the target and predicted future chunk correspondingly. Then, the training loss can be written as:
\begin{equation}
    \mathcal{L}_{\mathrm{pred}}
    =
    \mathrm{MSE}
    \left(
        \hat{\mathbf{Z}}_{t}^\mathrm{{tac}},
        \mathbf{Z}_{t}^\mathrm{{tac}}
    \right)
    +
    \lambda_{\mathrm{dyn}}
    \mathrm{MSE}
    \left(
        \nabla \hat{\mathbf{Z}}_{t}^\mathrm{{tac}},
        \nabla \mathbf{Z}_{t}^\mathrm{{tac}}
    \right),
\end{equation}
where $\nabla$ denotes the first-order temporal difference along the chunk dimension.

Inspired by LeJEPA~\cite{balestriero2025lejepa}, we incorporate the Sketched Isotropic Gaussian Regularizer (SIGReg) to avoid representation collapse during training. 
In our setting, SIGReg regularizes the tactile latent distribution toward an isotropic Gaussian structure. 
The final world-model objective is:
\begin{equation}
    \mathcal{L}_{\mathrm{WM}}
    =
    \mathcal{L}_{\mathrm{pred}}
    +
    \lambda_{\mathrm{sig}}
    \mathcal{L}_{\mathrm{sig}} .
\end{equation}

\subsection{Predictive Tactile-conditioned Policy}
To predict contact-aware action sequences, we design a predictive tactile-conditioned policy. Given multimodal observations, the policy extracts visual, proprioceptive, and tactile latent features, models the interaction between current and predicted tactile latents, and fuses the resulting tactile representation with visual features. 
A conditional flow-matching action head then predicts future action sequences conditioned on the fused multimodal features.

\subsubsection{Multimodal Feature Extraction}
The multimodal encoder extracts visual context, proprioceptive history, and tactile latent dynamics from raw observations.
Given the current RGB observation at time $t$, a frozen DINOv2-small~\cite{oquab2023dinov2} backbone encodes it into the visual feature $\mathbf{h}^{\mathrm{img}}_{t}$. 
The recent proprioceptive history $\mathbf{s}_{t-K+1:t}$ is flattened and encoded by an MLP into the state feature $\mathbf{h}^{\mathrm{s}}_{t}$. 
For tactile conditioning, the tactile tokenizer encodes recent H-frames tactile observations into the tactile latents
$\mathbf{z}_{t-H:t}^\mathrm{{tac}}$.
The pretrained world model then predicts future tactile latents $\hat{\mathbf{Z}}^{\mathrm{tac}}_{t}$ from $\mathbf{z}^\mathrm{{tac}}_{t-H:t}$ and aligned wrist force/torque features $\mathbf{c}_{t-H:t}$. 
These extracted multimodal features are used for policy conditioning.

\subsubsection{Current-Future Tactile Interaction}

While current tactile latents describe the present physical interaction states, predicted latents capture anticipatory dynamics. To explicitly model this interaction between current and future states, we introduce a cross-attention mechanism~\cite{vaswani2017attention}. Before cross-attention, learnable temporal embeddings are added to each latent to preserve the temporal order within each sequence:
\begin{equation}
    \bar{\mathbf{Z}}_{t, cur}^\mathrm{{tac}}
    =
    \mathbf{z}^\mathrm{{tac}}_{t-H:t}
    +
    \mathbf{E}_{temp},
    \quad
    \bar{\mathbf{Z}}_{t, fut}^\mathrm{{tac}}
    =
    \hat{\mathbf{Z}}_{t}^\mathrm{{tac}}
    +
    \mathbf{E}_{temp}.
\end{equation}

To enable the current tactile state to retrieve relevant future interaction cues, we employ the current tactile latents as queries, while utilizing the predicted future latents as keys and values. The cross-attention ($\mathrm{CA}$) is written as:
\begin{equation}
    \mathbf{H}_{t}^\mathrm{{tac}}
    =
    \bar{\mathbf{Z}}_{cur}^\mathrm{{tac}}
    +
    \mathrm{CA}
    \left(
        Q=\bar{\mathbf{Z}}_{t, cur}^\mathrm{{tac}},
        K, V=\bar{\mathbf{Z}}_{t, fut}^\mathrm{{tac}}
        % V=\bar{\mathbf{Z}}_{t, fut}^\mathrm{{tac}}
    \right).
\end{equation}
This residual structure augments the immediate contact state with predicted tactile dynamics. The enhanced tactile sequence $\mathbf{H}_{t}^\mathrm{{tac}}$ is then averaged over the temporal dimension to obtain a compact future-aware tactile representation $\mathbf{h}_{t}^\mathrm{{tac}}$, which is used for subsequent visuo-tactile fusion.

\subsubsection{Adaptive Visuo-Tactile Fusion}
Rather than directly concatenating visual and tactile features, we introduce an adaptive channel-wise visuo-tactile fusion module that uses predictive tactile representations to dynamically regulate modality contributions according to interaction dynamics.

In contrast to prior token-level visuo-tactile fusion approaches~\cite{zheng2026omnivta}, our method performs fusion at the feature-channel granularity. Conditioned on $\mathbf{h}_{t}^\mathrm{{tac}}$, this gate enables the policy to dynamically balance visual context and predictive tactile dynamics. Specifically, the tactile-guided gate is generated by an MLP with sigmoid activation:
\begin{equation}
    \boldsymbol{\alpha}
    =
    \sigma
    \left(
        \mathrm{MLP}
        \left(
            \mathbf{h}_{t}^\mathrm{{tac}}
        \right)
    \right).
\end{equation}

We then project the visual and tactile features into a shared space and perform channel-wise adaptive fusion as:
\begin{equation}
    \mathbf{h}_{t}^{\mathrm{vt}}
    =
    (\mathbf{1}-\boldsymbol{\alpha})
    \odot
    \mathbf{h}_{t}^{\mathrm{img}}
    +
    \boldsymbol{\alpha}
    \odot
    \mathbf{h}_{t}^{\mathrm{tac}}.
\end{equation}
Here, $\boldsymbol{\alpha}$ is computed from tactile representation and optimized with the policy objective.

\subsubsection{Flow-matching Action Head}
The fused visuo-tactile feature $\mathbf{h}_{t}^{\mathrm{vt}}$ and the proprioceptive feature $\mathbf{h}_{t}^{\mathrm{s}}$ are concatenated and passed through a condition encoder to form the global policy condition $\mathbf{y}_{t}$. Conditioned on $\mathbf{y}_{t}$, the action head predicts an action chunk
$\mathbf{A}_{t} = \mathbf{a}_{t:t+L-1} \in \mathbb{R}^{L \times d_a}$.
To model the conditional action distribution, we adopt a lightweight conditional flow matching framework~\cite{lipman2022flow}.

Let $\mathbf{A}_{t}^{(1)}$ denote the expert action chunk and
$\mathbf{A}_{t}^{(0)} \sim \mathcal{N}(\mathbf{0},\mathbf{I})$ denote a Gaussian noise action chunk.
Given a sampled flow time $\tau \sim \mathcal{U}(0,1)$, we construct the linear interpolation path:
\begin{equation}
    \mathbf{A}_{t}^{(\tau)}
    =
    (1-\tau)\mathbf{A}_{t}^{(0)}
    +
    \tau \mathbf{A}_{t}^{(1)},
    \quad
    \mathbf{u}_{t}
    =
    \mathbf{A}_{t}^{(1)}
    -
    \mathbf{A}_{t}^{(0)} .
\end{equation}

A temporal U-Net $v_{\theta}$ is trained to predict the conditional velocity field along this trajectory:
\begin{equation}
    \mathcal{L}_{\mathrm{FM}}
    =
    \mathbb{E}_{\mathbf{A}_{t}^{(0)}, \mathbf{A}_{t}^{(1)}, \tau}
    \left[
    \left\|
        v_{\theta}
        \left(
            \mathbf{A}_{t}^{(\tau)},
            \tau,
            \mathbf{y}_{t}
        \right)
        -
        \mathbf{u}_{t}
    \right\|_2^2
    \right].
\end{equation}

During inference, we initialize $\mathbf{A}_{t}^{0}\sim\mathcal{N}(\mathbf{0},\mathbf{I})$, and progressively refine by integrating the learned ordinary differential equation from $\tau=0$ to $\tau=1$ to obtain the denoised action chunk $\hat{\mathbf{A}}_{t}$:
\begin{equation}
\frac{d\mathbf{A}{t}^{(\tau)}}{d\tau}
=
v_{\theta}
\left(
\mathbf{A}{t}^{(\tau)},
\tau,
\mathbf{y}{t}
\right).
\end{equation}
\begin{figure*}[t]
    \centering
    \includegraphics[width=\textwidth]{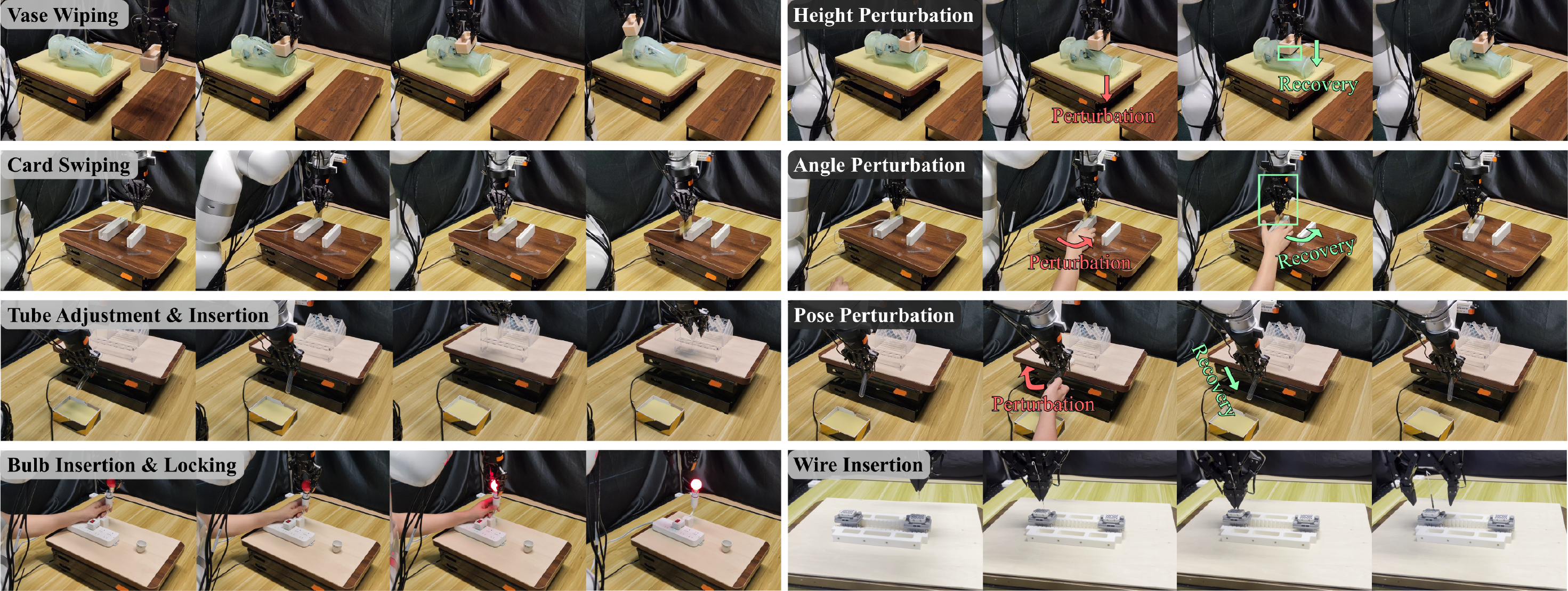}
    \vspace{-5mm}
    \caption{
    \textbf{Overview of the contact-rich manipulation tasks.}
    We evaluate policies on \textbf{five core contact-rich tasks}:
    Vase Wiping (Wiping), Card Swiping (Swiping),
    Tube Adjustment and Insertion (Adjustment), Bulb Insertion and Locking (Locking),
    and Wire Insertion (Insertion).
    We further introduce \textbf{three in-process dynamic perturbation settings}:
    height perturbation during wiping (Wiping-P), angle perturbation during swiping (Swiping-P),
    and pose perturbation during tube adjustment (Adjustment-P).
    }
    \label{fig:tasks}
\end{figure*}

\begin{table*}[!t]
\centering
\caption{
Manipulation performance on contact-rich manipulation tasks and in-process perturbation tasks.
}
\vspace{-2mm}
\label{tab:task_performance}
\small
\renewcommand{\arraystretch}{1.15}
\begin{tabularx}{0.95\textwidth}{@{}>{\raggedright\arraybackslash}p{1.7cm}*{8}{Y}@{}}
\toprule
\multicolumn{1}{c}{\multirow{2}{*}{\textbf{Policy}}}
& \multicolumn{5}{c}{\textbf{Contact-Rich Manipulation Tasks}} 
& \multicolumn{3}{c}{\textbf{In-Process Perturbations Tasks}} \\
\cmidrule(lr){2-6} \cmidrule(lr){7-9}
& \textbf{Wiping} 
& \textbf{Swiping} 
& \textbf{Adjustment} 
& \textbf{Locking} 
& \textbf{Insertion} 
& \textbf{Wiping} 
& \textbf{Swiping} 
& \textbf{Adjustment} \\
\midrule
DP~\cite{chi2025diffusion}        & 70\% & 35\% & 30\% & 10\% & 15\% & 0\%  & 0\%  & 0\%  \\
DP+Tactile+Force                  & 80\% & 40\% & 35\% & 30\% & 15\% & 25\% & 0\%  & 35\% \\
KineDex~\cite{zhang2025kinedex}   & 30\% & 35\% & 25\% & 45\% & 30\% & 10\% & 0\%  & 0\%  \\
FoAR~\cite{he2025foar}            & 50\% & 50\% & 35\% & 25\% & 20\% & 30\% & 0\%  & 25\% \\
RDP~\cite{xue2025reactive}        & 85\% & 50\% & 25\% & 55\% & 0\%  & 35\% & 65\% & 0\%  \\
\textbf{Ours}       & \textbf{100\%} & \textbf{85\%} & \textbf{70\%} & \textbf{80\%} 
                    & \textbf{60\%} & \textbf{90\%} & \textbf{85\%} & \textbf{85\%} \\
\bottomrule
\end{tabularx}
\end{table*}

\section{EXPERIMENTS}
% We evaluate the proposed method on a diverse set of contact-rich manipulation tasks and conduct ablation studies to quantify the contribution of each key component.
In this section, we first describe our experimental setup. 
We then evaluate the effectiveness of our method on diverse contact-rich manipulation tasks. Finally,
we conduct ablation studies to quantify the contribution of each key component. 

% set up
% main results: performance分析简单一段（总结）
% ablation studies (1) 触觉表征： 时序和空间  (2) condition+时序horizen 对表征影响  （3）component对操作影响
\subsection{Experimental Setup}

\subsubsection{Robotic Platform}
% Experiments were conducted on a UFactory xArm7 robotic arm equipped with a Robotiq two-finger gripper. An Intel RealSense D435 camera was mounted on the robot wrist to provide RGB observations at 30 Hz. For force-tactile sensing, the robot was equipped with a UFACTORY six-axis force/torque sensor, which recorded wrist interaction forces at 120 Hz, and two fingertip tactile sensors operating at 30 Hz. Each tactile sensor provides reconstructed 3D displacement fields, which are represented as 35×2035 \times 20 spatial tactile maps.
To evaluate our method on a real robotic platform, we use a 7-DoF UFactory xArm7 robot equipped with a Robotiq 2F-85 gripper, an Intel RealSense D435 wrist-mounted camera, a UFactory 6-axis force/torque sensor, and two Xense tactile sensors equipped on each fingertip. Image and tactile observations are captured at 30 Hz, while wrist wrench data are recorded at 120 Hz. Each tactile sensor outputs a $35 \times 20$ 3D displacement map representing the local contact deformation.

\subsubsection{Training Details}
The Predictive Force-conditioned Tactile World Model (11.8M parameters) is trained on 2,700 force-tactile interaction episodes, including both task-specific demonstrations and diverse contact interaction data. The model is optimized for 150k steps with $\lambda_{\mathrm{dyn}} = 1.00$ and $\lambda_{\mathrm{sig}} = 0.09$. We then freeze the pretrained tactile encoder and predictor to extract latent tactile representations for the downstream flow-matching policy (68.9M parameters). All experiments are trained on 8 NVIDIA A100 GPUs.

\subsubsection{Baselines}
We compare our method against several representative imitation learning baselines:
\begin{itemize}
    \item \textbf{DP}~\cite{chi2025diffusion}: a visuomotor imitation policy conditioned on wrist RGB observations and proprioceptive states.
    
    \item \textbf{DP+Tactile+Force}: a multimodal extension of DP that additionally incorporates tactile features and wrist wrench observations to predict action sequences.
    
    \item \textbf{KineDex}~\cite{zhang2025kinedex}: a tactile-aware visuomotor policy that jointly predicts action sequences and tactile representations from current visual and tactile observations.
    
    \item \textbf{FoAR}~\cite{he2025foar}: a force-aware policy that modulates force representations using a future contact predictor. We replace the original point-cloud encoder with a 2D RGB encoder for fair comparison.
    
    \item \textbf{RDP}~\cite{xue2025reactive}: a reactive visual-tactile policy that leverages tactile and force feedback for online action refinement.
\end{itemize}

% \subsection{Evaluation Protocol}
% As shown in Fig.~\ref{fig:tasks}, our benchmark consists of two categories: contact-rich task execution and dynamics perturbed in-process manipulation. For perturbation settings, we collect additional recovery demonstrations in which the demonstrator re-establishes stable contact after disturbances and completes the task. For each method and each task, we conduct 20 independent real-world trials and report the average task completion score. For Wiping and Swiping, the score is computed as the ratio of the successfully completed wiping or swiping length to the target length. 
% For Tube Adjustment and Bulb Locking, which involve two sequential stages, completing one stage receives 50\% and completing both receives 100\%. 
% For Wire Insertion, success requires inserting the wire into the target socket. 
% For perturbation settings, a trial is successful if the policy recovers from the disturbance and completes the task.

\begin{figure*}[t]
    \centering
    \includegraphics[width=\textwidth]{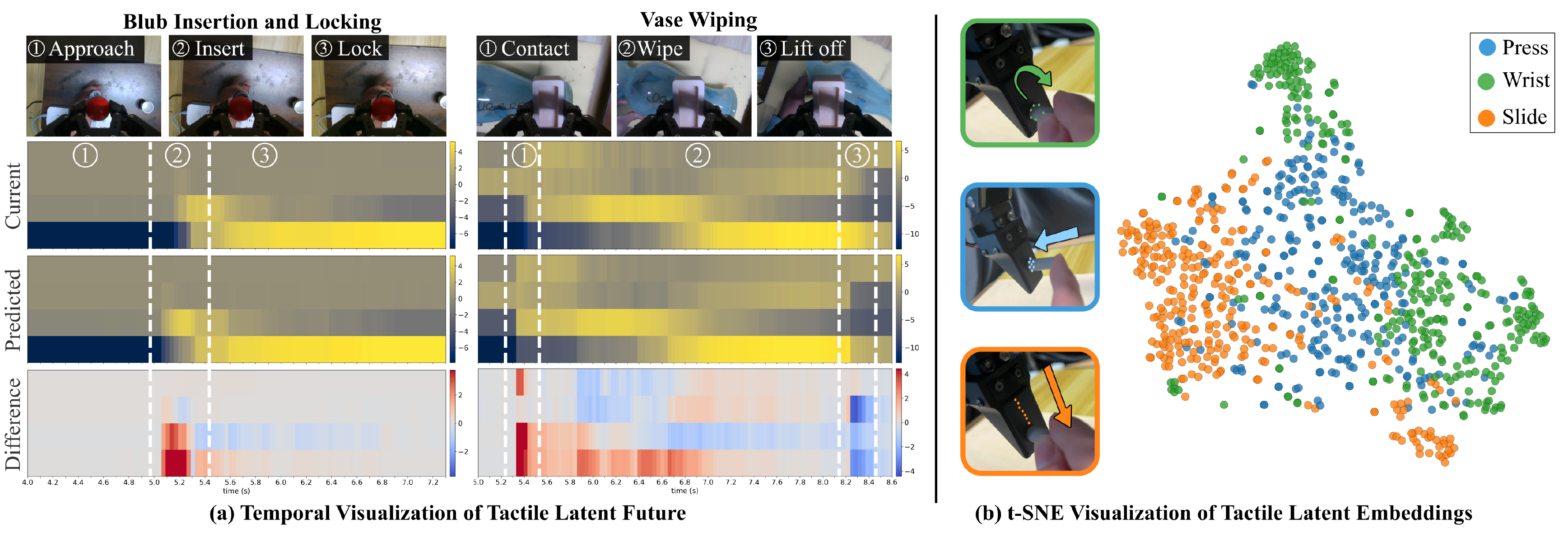}
    \vspace{-8mm}
    \caption{
    \textbf{Tactile latent representation analysis.}
    (a) Temporal visualization of tactile latents during Bulb Insertion and Locking and Vase Wiping.  
    (b) t-SNE visualization of tactile latent embeddings on different primitive interactions.
    }
    \label{fig:latent_feature}
    \vspace{-3mm}
\end{figure*}

\subsection{Benchmark Tasks}

\subsubsection{Contact-Rich Manipulation Tasks}

As shown in Fig.~\ref{fig:tasks}, we evaluate our method on five representative contact-rich manipulation tasks: \textit{Vase Wiping}, \textit{Card Swiping}, \textit{Tube Adjustment \& Insertion}, \textit{Bulb Insertion \& Locking}, and \textit{Wire Insertion}. \textit{Wiping} and \textit{Swiping} require maintaining stable sliding contact during long-horizon surface interactions. \textit{Tube Adjustment} and \textit{Bulb Locking} involve sequential alignment and constrained contact manipulation. \textit{Wire Insertion} requires precise contact control to insert a flexible wire into a socket.

\subsubsection{Perturbation-Aware Evaluation}

We evaluate all methods under two settings: standard contact-rich task execution and perturbation-aware manipulation. In the standard setting, policies are trained using task-specific demonstrations collected under nominal execution conditions. In the perturbation-aware setting, we additionally collect recovery demonstrations in which external disturbances are introduced during task execution, requiring the demonstrator to re-establish stable contact and recover the manipulation process. Policies in this setting are trained using both nominal demonstrations and recovery interaction data.

\subsubsection{Evaluation Metrics}

For each method and each task, we conduct 20 independent trials and report the average task completion score. For \textit{Wiping} and \textit{Swiping}, the score is defined as the ratio between the completed interaction length and the target length. For \textit{Tube Adjustment} and \textit{Bulb Locking}, which consist of two sequential stages, completing the first stage yields a score of 50\%, while completing both stages yields 100\%. For Wire Insertion, a trial is considered successful only if the wire is fully inserted into the target socket. Under perturbation-aware evaluation, a trial is counted as successful only if the policy successfully recovers from the disturbance and completes the task.

\subsection{Main Results}
\subsubsection{Real-Robot Task Performance}
We present quantitative performance in Table~\ref{tab:task_performance} and execution sequences in Fig.~\ref{fig:tasks}. 
The results show that our method achieves the best performance across both nominal contact-rich manipulation tasks and in-process perturbation settings, demonstrating the effectiveness of predictive tactile-latent priors for robust contact reasoning and control.

On the five representative contact-rich manipulation tasks, our method achieves an average completion score of 79.0\%, substantially outperforming all baselines. 
While vision-only and reactive feedback-based policies can handle relatively simple or sustained-contact interactions, their performance drops on tasks requiring precise contact regulation and multi-stage contact transitions. 
By contrast, our method consistently improves performance across all tasks by leveraging predictive tactile-latent priors, which provide anticipatory contact information for more reliable contact establishment, maintenance, and task completion.

This advantage becomes even more evident under in-process perturbations. 
Across height, angle, and pose disturbances, our method achieves 90\%, 85\%, and 85\%, respectively, with an average score of 86.7\%, clearly outperforming all baselines. 
These results show that predictive tactile-latent priors improve recovery robustness when contact conditions change dynamically during execution.

Beyond the quantitative results, the qualitative sequences in Fig.~\ref{fig:tasks} provide representative examples of perturbation recovery. 
After height, angle, and pose disturbances, our policy can re-establish stable contact, correct the end-effector motion, and continue the task without losing progress. 

\subsubsection{Predictive Tactile Representation Analysis}
We further analyze the learned tactile latent representations from two perspectives: temporal prediction and representation discriminability. 
First, we examine whether the tactile world model produces anticipatory latent responses around contact-state transitions. 
Second, we evaluate whether the tactile spatial encoder learns discriminative representations for different force--tactile interaction patterns.

For temporal analysis, we visualize tactile latents during Bulb Insertion and Locking and Vase Wiping. 
The current tactile latent, predicted future tactile latent, and their latent difference are projected into a low-dimensional PCA space for visualization. 
As shown in Fig.~\ref{fig:latent_feature}(a), the predicted tactile latents exhibit contact-related variations approximately 200 ms before similar changes appear in the current tactile latents. 
The latent difference responses become prominent around key contact-stage transitions, suggesting that the world model captures temporal contact evolution and provides anticipatory contact cues.

We also evaluate representation generalization on unseen force--tactile interaction episodes, including pressing, twisting, and sliding. 
For each episode, frame-level tactile latents are extracted by the pretrained tactile encoder and aggregated into sequence-level embeddings using temporal max pooling. 
The t-SNE visualization in Fig.~\ref{fig:latent_feature}(b) shows that the embeddings form well-separated clusters according to the underlying contact pattern. 
This indicates that the tactile encoder learns contact-discriminative representations that capture local deformation and force-induced interaction changes beyond the training distribution.

\subsection{Ablation Studies}
We conduct ablation studies along two aspects. 
The first aspect investigates the conditioning design of the tactile world model, where we compare different modality combinations to determine which physical cues are most effective for forecasting future tactile latents. 
The second aspect evaluates the role of force-tactile integration in policy learning, focusing on how different fusion strategies and tactile usage schemes influence downstream manipulation performance. 

\subsubsection{World Model Conditioning}
We first ablate the conditioning modality used in the tactile world model.
We compare four conditioning designs: no external condition, RGB image condition, robot state condition, and wrist wrench condition. 
Prediction quality is evaluated using Mean Squared Error (MSE), cosine similarity (Cos), and symmetric KL divergence (KL$_{\mathrm{sym}}$), which respectively measure element-wise prediction error, latent directional consistency, and latent distribution consistency.
As shown in Table~\ref{tab:condition_ablation}, wrist wrench conditioning achieves the best performance across all metrics. 
Compared with the no-condition variant, it reduces MSE from 0.027 to 0.017 and KL$_{\mathrm{sym}}$ from 0.014 to 0.009, while increasing cosine similarity from 0.954 to 0.992. 
These results indicate that wrist wrench provides the most informative physical condition for forecasting tactile contact evolution, as it directly captures physical interaction changes during contact-rich manipulation.

\begin{table}[!t]
\centering
\caption{
Ablation of conditioning modality for tactile world model.
}
\vspace{-2mm}
\label{tab:condition_ablation}
\small
\renewcommand{\arraystretch}{1.12}
\begin{tabular*}{0.85\columnwidth}{@{\extracolsep{\fill}}lccc@{}}
\toprule
\textbf{Condition} 
& \textbf{MSE}$\downarrow$ 
& \textbf{Cos}$\uparrow$
& \textbf{KL$_{\mathrm{sym}}$}$\downarrow$ \\
\midrule
w/o Condition & 0.027 & 0.954 & 0.014 \\
RGB Image       & 0.025 & 0.973 & 0.013 \\
% Delta Force  & 0.0256 & 0.9885 & 0.0132 \\
Robot State      & 0.022 & 0.975 & 0.011 \\
Wrist Wrench        & \textbf{0.017} & \textbf{0.992} & \textbf{0.009} \\
\bottomrule
\end{tabular*}
\end{table}

\begin{table}[t]
\centering
\caption{
Ablation of the predictive tactile policy.
}
\vspace{-2mm}
\label{tab:policy_ablation}
\small
\setlength{\tabcolsep}{4pt}
\renewcommand{\arraystretch}{1.12}
\begin{tabular}{lccc}
\toprule
\textbf{Method} 
& \textbf{Wiping} 
& \textbf{Wiping-P} 
& \textbf{Swiping-P} \\
\midrule
% \multicolumn{4}{l}{\textit{Component ablation}} \\
Parallel fusion            & 80\%  & 25\% & 10\% \\
w/o force condition        & 70\%  & 50\% & 75\% \\
w/o predicted tactile      & 65\%  & 15\% & 15\% \\
w/o cross-attention        & 100\% & 65\% & 0\%  \\
w/o adaptive gate          & 90\%  & 65\% & 75\% \\
\textbf{Ours}              & \textbf{100\%} & \textbf{85\%} & \textbf{90\%} \\
% \midrule
% \multicolumn{4}{l}{\textit{Tactile-prior noise ablation}} \\
% 25\% Noise                 & 100\% & 80\% & 90\% \\
% 50\% Noise                 & 100\% & 65\% & 90\% \\
% 75\% Noise                 & 100\% & 10\% & 80\% \\
% 100\% Noise                & 80\%  & 0\%  & 60\% \\
\bottomrule
\vspace{-7mm}
\end{tabular}
\end{table}

% We further analyze the effect of prediction horizon on tactile latent forecasting. 
% Since tactile observations are recorded at 30 Hz, future offsets of 8, 12, 16, 20, and 24 frames correspond to approximately 267 ms, 400 ms, 533 ms, 667 ms, and 800 ms, respectively. 
% As shown in Fig.~\ref{fig:prediction}, the prediction error increases as the horizon becomes longer, with MSE rising from 0.0179 at 267 ms to 0.0333 at 800 ms. 
% This trend indicates that long-horizon tactile forecasting ij our setting becomes increasingly uncertain due to complex contact evolution. 

% \begin{figure}[h]
%     \centering
%     \includegraphics[width=0.8\columnwidth]{figs/prediction.pdf}
%     \vspace{-2mm}
%     \caption{
%     \textbf{Effect of prediction horizon on tactile latent forecasting.}
%     }
%     \label{fig:prediction}
% \end{figure}

\begin{figure}[t]
    \centering
    \includegraphics[width=0.95\columnwidth]{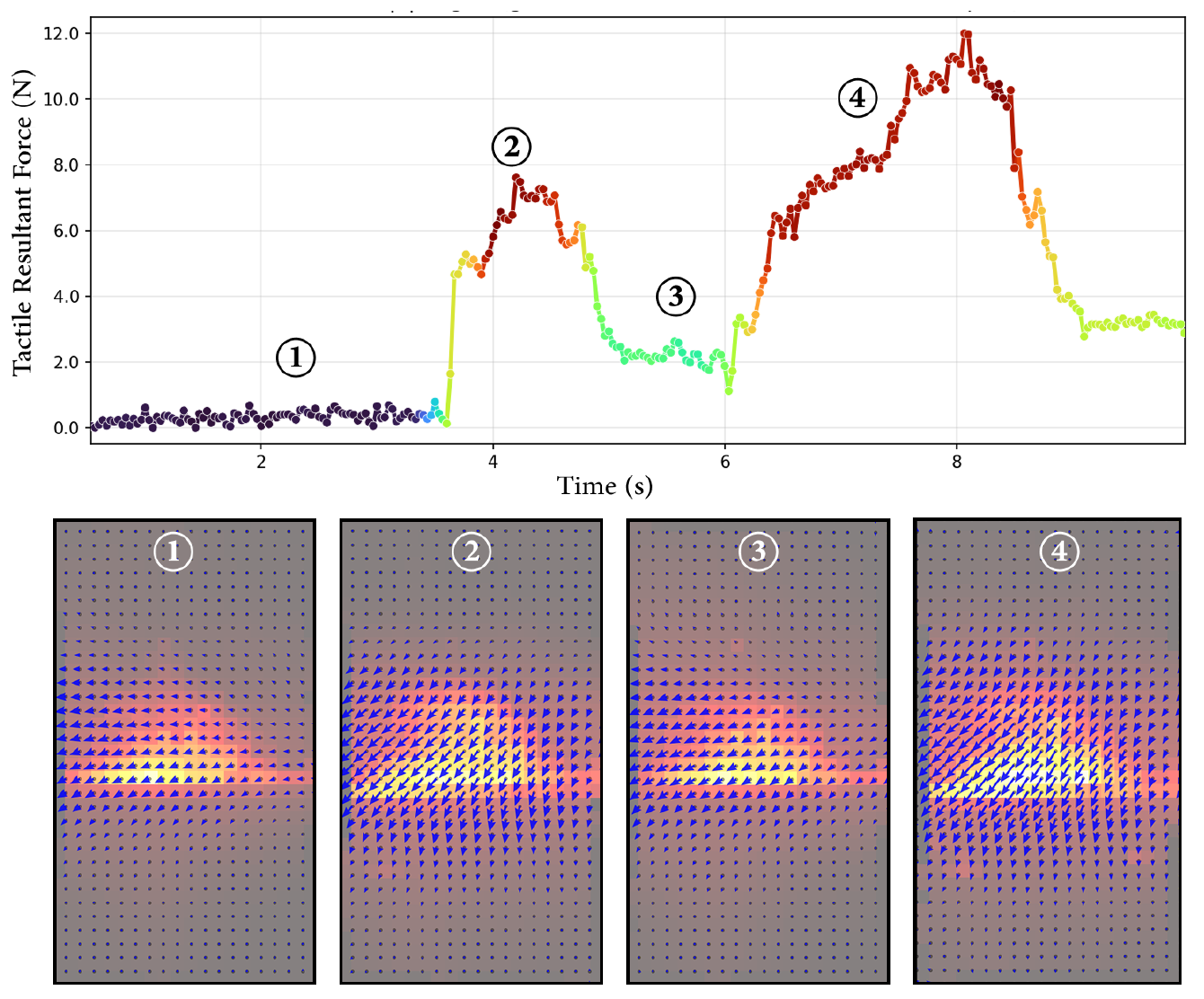}
    \vspace{-2mm}
    \caption{
    \textbf{Visualization of tactile gating on the Vase Wiping Perturbation.}
    The top panel shows the tactile resultant force trajectory, with gating features projected to one dimension by PCA and overlaid as a colormap.
    The bottom panel shows representative tactile observations from four interaction stages.
    }
    \label{fig:gate_visualization}
    \vspace{-5mm}
\end{figure}

\subsubsection{Policy Ablation}
We next ablate how tactile information is used and integrated into the policy, focusing on the roles of force conditioning, predicted tactile feature, current--future tactile interaction, and adaptive visuo-tactile fusion. 
We compare the following policy variants:
\begin{itemize}
    \item \textbf{Parallel Fusion}: directly fuses tactile features and wrist wrench features as policy input.
    \item \textbf{w/o Force Condition}: removes wrist wrench conditioning from the predictive tactile prior.
    \item \textbf{w/o Predicted Tactile}: removes predicted tactile features and uses only current tactile features as input.
    \item \textbf{w/o Cross-attention}: replaces current--future tactile cross-attention with direct feature concatenation.
    \item \textbf{w/o Adaptive Gate}: replaces adaptive visuo-tactile gating with direct feature concatenation.
\end{itemize}

For each variant, we conduct 20 real-world trials under the same evaluation protocol and report the average task completion score. 
As shown in Table~\ref{tab:policy_ablation}, the full model achieves the best performance across all settings, with 100\% on Wiping, 85\% on Wiping-P, and 90\% on Swiping-P. 
In contrast, \textit{Parallel Fusion} performs poorly under perturbations, showing that simple force--tactile concatenation is insufficient for robust contact recovery. 
Removing predicted tactile features causes a clear drop on both Wiping-P and Swiping-P, indicating the importance of future tactile priors for maintaining and recovering contact. 
The degradation of \textit{w/o Force Condition} and \textit{w/o Cross-attention} further confirms that wrist-wrench conditioning and explicit current--future tactile interaction are both important for effectively using predicted tactile.

Removing the adaptive gate also reduces Wiping-P performance and increases recovery time from 2.56~s to 4.06~s, suggesting that adaptive visuo-tactile fusion improves contact regulation during perturbation recovery. 
To interpret this effect, Fig.~\ref{fig:gate_visualization} visualizes the learned tactile-guided gate by projecting the channel-wise gate to one dimension with PCA and overlaying it on the tactile resultant force trajectory. 
The gate responses vary across contact stages, indicating that the adaptive gate modulates tactile information according to changing contact conditions.

\section{Conclusion}
% In this paper, we presented \textbf{TacForeSight}, a lightweight force-conditioned tactile foresight framework for contact-rich manipulation. 
% By predicting future tactile latents conditioned on high-rate wrist force/torque signals, TacForeSight provides anticipatory contact priors for downstream action generation. 
% The proposed predictive tactile-conditioned policy further models current--future tactile evolution and adaptively fuses visual and tactile features for robust control.
% Real-robot experiments on five contact-rich tasks and three perturbation settings demonstrate that TacForeSight outperforms SOTA baselines, especially under dynamic contact changes. 

In this paper, we introduce \textbf{TacForeSight}, which moves beyond previous reactive feature fusion to establish a proactive and predictive paradigm for contact-rich manipulation.
By conditioning tactile latent prediction on wrist force/torque signals, TacForeSight captures the asymmetric relationship between global force/torque changes and local tactile evolution.
Real-robot experiments across contact-rich tasks and dynamic perturbations demonstrate its effectiveness for robust contact regulation and disturbance recovery. 
These results indicate that latent-space tactile prediction is an effective and efficient mechanism for bringing predictive physical reasoning into real-time contact-rich manipulation.

% \section*{APPENDIX}
% \input{sections/appendix}

% \section*{ACKNOWLEDGMENT}
% \input{sections/ack}

\bibliographystyle{IEEEtran}
\bibliography{ref}

\end{document}